%% file: LoGoSeg.tex
\documentclass[letterpaper]{article} 
\usepackage{aaai2026}  
\usepackage{times}  
\usepackage{helvet}  
\usepackage{courier}  
\usepackage[hyphens]{url}  
\usepackage{graphicx} 
\usepackage{natbib}  
\usepackage{caption} 
\usepackage{algorithm}
\usepackage{algorithmic}

\usepackage{newfloat}
\usepackage{listings}
\usepackage{booktabs}
\usepackage{amsfonts} 
\usepackage{amssymb}

\usepackage{multirow}
\usepackage[table]{xcolor}
\usepackage{colortbl}  
\usepackage{pifont}    
\newcommand{\xmark}{\ding{55}}  
\newcommand{\cmark}{\ding{51}}  
\usepackage{amsmath}

\definecolor{topColor}{RGB}{247,252,253}
\definecolor{middleColor}{RGB}{247,252,245} 
\definecolor{bottomColor}{RGB}{255,255,229}  

\usepackage{newfloat}
\usepackage{listings}
\DeclareCaptionStyle{ruled}{labelfont=normalfont,labelsep=colon,strut=off} 
\floatstyle{ruled}
\newfloat{listing}{tb}{lst}{}
\floatname{listing}{Listing}
\title{LoGoSeg: Integrating Local and Global Features for Open-Vocabulary Semantic Segmentation}
\author{
    Junyang Chen\textsuperscript{\rm 1,2},
    Xiangbo Lv\textsuperscript{\rm 1,2,3},
    Zhiqiang Kou\textsuperscript{\rm 1,2},
    Xingdong Sheng\textsuperscript{\rm 3},
    Ning Xu\textsuperscript{\rm 1,2},
    Yiguo Qiao\textsuperscript{\rm 1,2}\thanks{Corresponding author.}
}

\affiliations{
    \textsuperscript{\rm 1}School of Computer Science and Engineering, Southeast University, Nanjing 211189, China\\
    \textsuperscript{\rm 2}Key Laboratory of Computer Network and Information Integration (Ministry of Education), Southeast University, Nanjing 211189, China\\
    \textsuperscript{\rm 3}Lenovo Research, Shanghai 201203, China\\
    \{chenjunyang, zhiqiang\_kou, xning, yqiao\}@seu.edu.cn, \{lvxb6, shengxd1\}@lenovo.com
}

\usepackage{bibentry}

\begin{document}

\maketitle

\begin{abstract}
Open-vocabulary semantic segmentation (OVSS) extends traditional closed-set segmentation by enabling pixel-wise annotation for both seen and unseen categories using arbitrary textual descriptions. While existing methods leverage vision-language models (VLMs) like CLIP, their reliance on image-level pretraining often results in imprecise spatial alignment, leading to mismatched segmentations in ambiguous or cluttered scenes. However, most existing approaches lack strong object priors and region-level constraints, which can lead to object hallucination or missed detections, further degrading performance. To address these challenges, we propose LoGoSeg, an efficient single-stage framework that integrates three key innovations: (i) an object existence prior that dynamically weights relevant categories through global image-text similarity, effectively reducing hallucinations; (ii) a region-aware alignment module that establishes precise region-level visual-textual correspondences; and (iii) a dual-stream fusion mechanism that optimally combines local structural information with global semantic context.  Unlike prior works, LoGoSeg eliminates the need for external mask proposals, additional backbones, or extra datasets, ensuring efficiency.  Extensive experiments on six benchmarks (A-847, PC-459, A-150, PC-59, PAS-20, and PAS-20$^b$) demonstrate its competitive performance and strong generalization in open-vocabulary settings.
\end{abstract}


\begin{figure}[!t]
\centering
\includegraphics[width=1.0\linewidth]{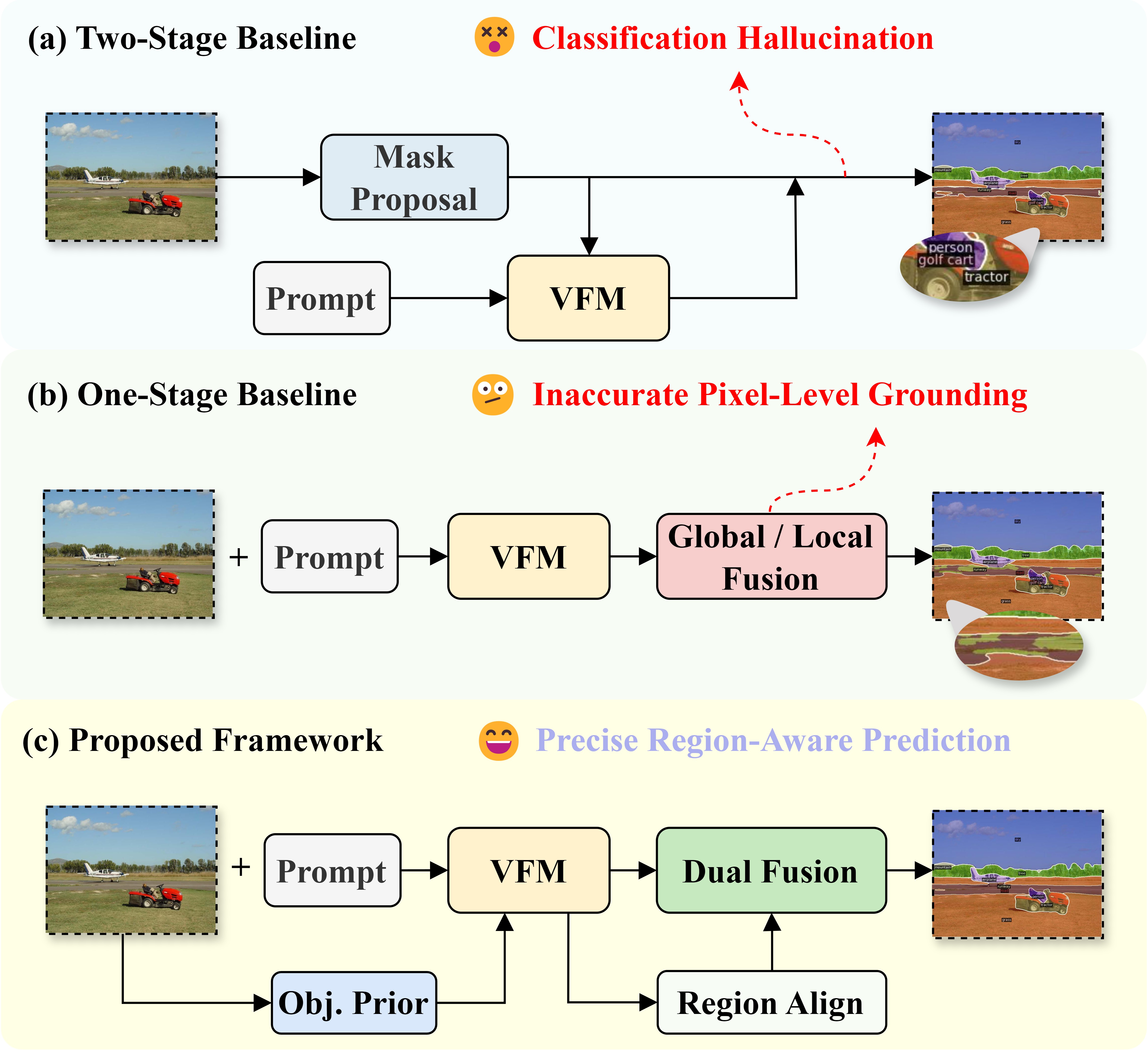}
\caption{
Comparison of open-vocabulary segmentation frameworks.
(Top) Two-stage methods rely on external mask proposals, often causing hallucinations.
(Middle) One-stage methods are more efficient but struggle with pixel-level grounding.
(Bottom) LoGoSeg integrates object priors, region alignment, and dual fusion to improve cross-modal consistency and segmentation quality.
}
\label{figure1}
\end{figure}

\section{Introduction}

Semantic segmentation, a fundamental computer vision task that assigns semantic labels to each image pixel, achieves high accuracy on fixed-category benchmarks \cite{mask2former, chen2018encoder, srivastava2024omnivec2}. However, these conventional methods suffer from overfitting, are constrained by predefined label sets, and fail to generalize to unseen classes. To address these limitations, open-vocabulary semantic segmentation (OVSS) has recently emerged, enabling pixel-level prediction for arbitrary categories specified via textual prompts \cite{zhao2017open, xian2019semantic, luddecke2022image}. This paradigm significantly broadens the applicability of semantic segmentation beyond closed-set constraints.

The core challenge of OVSS lies in its requirement to handle open-domain categories during inference—a capability fundamentally absent in traditional vision-only backbone networks. To address this, current research predominantly employs vision-language models (VLMs) such as CLIP \cite{clip}, ALIGN \cite{jia2021scaling}, EVA-CLIP \cite{sun2023eva} and the SigLIP series \cite{tschannen2025siglip, zhai2023sigmoid}, leveraging their zero-shot recognition capabilities acquired through large-scale image-text pretraining \cite{sun2025cliper, yang2025resclip}. However, these models' image-level supervision paradigm creates inherent difficulties in adapting to pixel-level segmentation tasks \cite{zhou2022extract, zhou2019semantic}. Specifically, the pretraining process lacks explicit constraints for aligning textual concepts with visual regions, resulting in blurred segmentation boundaries and uncertain category predictions—limitations that become particularly pronounced when dealing with unseen or rare categories in open-vocabulary scenarios. Furthermore, the absence of object priors and region-aware constraints may lead to false positives or missed detections in complex scenes, significantly compromising segmentation accuracy.

To address these challenges, researchers have developed two distinct methodological frameworks: two-stage \cite{liang2023open, ding2022open, Xu_2023_ICCV, ding2022decoupling, xu2022simple, ghiasi2022scaling, xu2023open, yu2023convolutions, jiao2024collaborative, wang2025dual} and single-stage \cite{dong2023maskclip, cho2024cat, zhou2023zegclip, xu2023san, xie2024sed, shan2024open, sun2025cliper} approaches. The two-stage paradigm, exemplified by mask proposal generators such as SAM \cite{kirillov2023segment, ravi2024sam, zhang2025corrclip} and MaskFormer \cite{mask2former}, predict category-agnostic mask proposals and then classify them using the zero-shot capability of frozen CLIP models. While achieving competitive accuracy, the efficiency is hindered by separate mask generation and classification stages. As shown in Fig.~\ref{figure1}(a), they fail to fully exploit contextual information and depend heavily on mask quality, which limits their generalization.  

Single-stage approaches, by contrast, directly adapt VLMs for pixel-level segmentation,  simultaneously enhancing efficiency while leveraging the models' inherent segmentation capabilities. While recent advances like SAN \cite{xu2023san}, SED \cite{xie2024sed}, and CAT-Seg \cite{cho2024cat} have demonstrated promising results,  Fig.~\ref{figure1}(b)  reveals   critical limitations:  an imbalance between local and global feature integration, insufficient region-level alignment, and unaddressed classification hallucinations.  

To overcome the limitations of current methods, we propose LoGoSeg, a unified single-stage framework that addresses three core challenges in OVSS: classification hallucinations, misaligned pixel-level supervision, and insufficient fusion of local and global cues. The proposed architecture incorporates three novel components:
first, an adaptive object existence prior computed from global image-text similarity dynamically adjusts category weights to suppress false positives; second, our region-aware alignment module establishes precise correspondences between visual regions and textual concepts through localized similarity computation; third, a dual-stream fusion mechanism jointly encodes fine-grained spatial structures and high-level semantic context in a unified representation space. Our contributions are summarized below:

\textbf{(i) Region-aware alignment with object prior}: Our strategy significantly reduces category hallucinations while improving pixel-level text-visual alignment through explicit region-level constraints.

\textbf{(ii) Dual-stream feature fusion}: The proposed framework effectively combines detailed local spatial context with comprehensive global semantics, overcoming the limitations of previous single-stream approaches.

\textbf{(iii) Unified single-stage framework}: LoGoSeg integrates these innovations into a single-stage pipeline that demonstrates superior performance across six standard benchmarks, achieving state-of-the-art results in both accuracy and computational efficiency.


\begin{figure*}[!t]
\centering
\includegraphics[width=\textwidth]{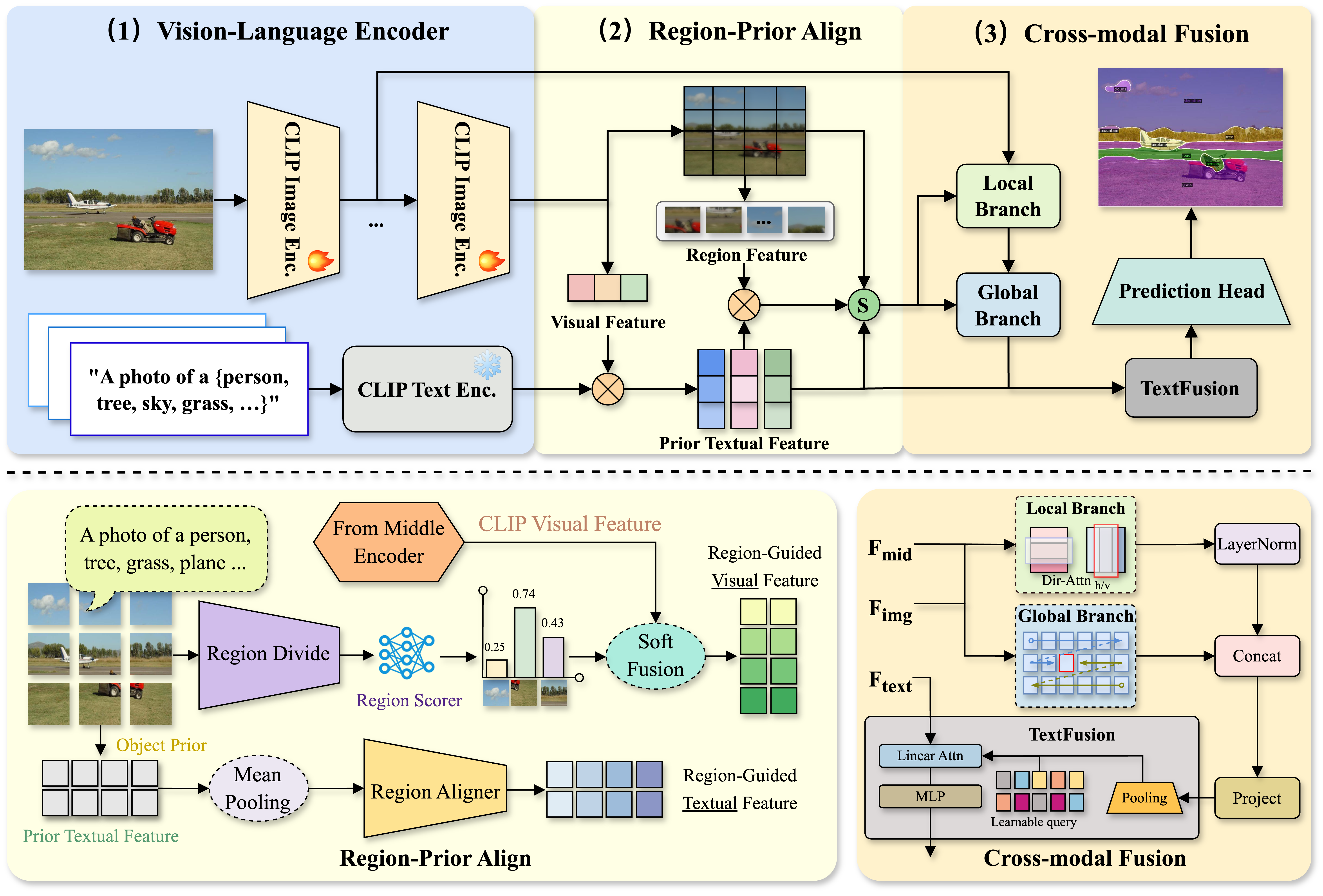}

\caption{Overview of LoGoSeg. CLIP encoders extract multi-level visual and textual embeddings. A prior-guided alignment module uses a lightweight MLP to score regions for category-aware guidance and a Region Aligner to tightly align visual and textual features. A dual-branch fusion integrates local directional self-attention with global state-space modeling. A transformer with learnable queries then performs fine-grained cross-modal fusion via linear attention, and a hierarchical decoder with learnable upsampling and guidance-modulated convolutions yields the final segmentation map.}

\label{model}
\end{figure*}

\section{Method}
We propose LoGoSeg, a region-aware framework for OVSS, as illustrated in Fig.~\ref{model}. Given an input image and a set of category prompts, the model first extracts image-level visual and textual features using CLIP. The prior-guided regional alignment module (\textbf{Section~\ref{sec:alignment}}) then measures similarity between regional visual features and adjusted text embeddings to provide category-aware guidance. These aligned features are subsequently passed to the cross-modal fusion module (\textbf{Section~\ref{sec:fusion}}), which integrates local and global cues via parallel architectures capturing directional spatial details and long-range contextual dependencies. Finally, the transformer-based decoder (\textbf{Section~\ref{sec:decoder}}) aggregates the fused features and predicts the final segmentation masks using learnable queries. Comprehensive details of each module are provided in the following subsections.

\subsection{Preliminary}
Open‐vocabulary semantic segmentation (OVSS) performs dense prediction of text-defined categories at the pixel level. Given an input image $\mathbf I \in\mathbb R^{H\times W\times3}$ and a set of N text prompts
$\mathcal T=\{\mathbf t_1,\dots,\mathbf t_N\}$, the task outputs a segmentation map 
$\mathbf Y\in\{1,\dots,N\}^{H\times W}$, where each pixel is assigned the index of its corresponding prompt. We use CLIP as the backbone, with image encoder $f_{\mathrm{img}}(\cdot)$ producing  
$\mathbf V=f_{\mathrm{img}}(\mathbf I)\in\mathbb R^{C\times H\times W}$ and text encoder  
$f_{\mathrm{text}}(\cdot)$ producing prompt embeddings  
$\mathbf T=f_{\mathrm{text}}(\mathcal T)\in\mathbb R^{N\times P\times C}$. However,
as CLIP is trained only at the image level, these embeddings lack direct pixel correspondence.  

\paragraph{Notation.}  Let $B$ denote batch size, $C$ feature channels, $H\times W$ spatial dimensions, $N$ the number of categories, $P$ prompts per category, $K$ number of regions, $D$ guidance dimension, $T$ number of queries, and $L_t$  pooled text tokens. The text embedding tensor is denoted as boldface $\mathbf T$, while the prompt set is calligraphic $\mathcal T$. We use $\|\cdot\|$ for $\ell_2$ norm, $[a\parallel b]$ for channel‐wise concatenation, and distinguish $\mathrm{softmax}_k$ (normalization over regions) from $\mathrm{softmax}_n$ (normalization over categories), respectively.

\subsection{Prior-Guided Regional Alignment}
\label{sec:alignment}

To improve category discriminability and reduce semantic confusion under open-vocabulary settings, we propose a prior-guided regional alignment module that integrates object priors with region-level vision-language correlation.


\paragraph{Object Prior Estimation.} Let $\mathbf{V} \in \mathbb{R}^{C \times H \times W}$ denote the CLIP-extracted image feature map, and $\mathbf{T} \in \mathbb{R}^{N \times P \times C}$ represent the prompt-based semantic text embeddings for $N$ categories with $P$ diverse prompts each. We compute the average prompt representation for each category as $\bar{\mathbf{T}}_n = \frac{1}{P} \sum_{p=1}^P \mathbf{T}_{n,p}$. The object-level prior is estimated by measuring the similarity between image features and the corresponding text embedding:
\begin{equation}
p_n^{\text{prior}} = \frac{1}{HW} \sum_{i=1}^{H} \sum_{j=1}^{W} \frac{\langle \mathbf{V}_{:,i,j}, \bar{\mathbf{T}}_n \rangle}{\|\mathbf{V}_{:,i,j}\| \cdot \|\bar{\mathbf{T}}_n\|}.
\end{equation}
This score quantifies the likelihood of category $n$, serving as a $\lambda$-scaled semantic prior. The weighted prompt center is then derived by reweighting the embeddings:
\begin{equation}
\hat{\bar{\mathbf T}}_n
= \frac{1}{P}\sum_{p=1}^{P} p_n^{\mathrm{prior}}\;\mathbf T_{n,p}.
\end{equation}


\paragraph{Region-Level Textual Guidance.}  
The feature map is partitioned into $K$ non-overlapping regions $\{R_k\}$, where $\mathbf{v}_k$ is mean-pooled within $R_k$. Given the $\lambda$-weighted prompt center $\hat{\bar{\mathbf{T}}}_n$, the region-category similarity matrix is defined as:
\begin{equation}
A_{k,n} = \frac{\langle \mathbf{v}_k, \hat{\bar{\mathbf{T}}}_n \rangle}{\|\mathbf{v}_k\| \cdot \|\hat{\bar{\mathbf{T}}}_n\|}, \quad A \in \mathbb{R}^{K \times N}.
\end{equation}

This similarity matrix is transformed into region-aware weights via temperature-scaled softmax normalization: 
\begin{equation}
w_{k,n} = \mathrm{softmax}_{k}\bigl(\tau\,A_{k,n}\bigr), 
\quad \sum_{k=1}^{K} w_{k,n}=1,
\end{equation}
where \(\tau\) adjusts the sharpness and \(w_{k,n}\) quantifies the contribution of region \(k\) to category \(n\). The visual prototype for category \(n\) is then obtained as:  
\begin{equation}
\mathbf{m}_n = \sum_{k=1}^{K} w_{k,n}\,\mathbf{v}_k, 
\quad \mathbf{m}_n \in \mathbb{R}^{C},
\end{equation}
where \(\mathbf{m}_n\) represents the aggregated visual feature for category \(n\). Finally, we form the region-level textual guidance by projecting the concatenation of the prototype and the weighted text center:
\begin{equation}
g_{\text{region}}[n] = W_t\,\bigl[\mathbf{m}_n \parallel \hat{\bar{\mathbf T}}_n\bigr] + \mathbf{b}_t,
\quad g_{\text{region}}[n] \in \mathbb{R}^{D}.
\end{equation}

\paragraph{Region-Level Visual Guidance.}  
For each region \(R_k\), a trainable two-layer MLP (Linear-GELU-Linear)  generates spatial attention logits, which are normalized via softmax to produce aggregation weights. These weights pool visual tokens into region embeddings \(g_{\text{image}}^{(k)}\). An optional linear projection then aligns its dimension to that of \(g_{\text{region}}\) for seamless cross-modal fusion.  


\paragraph{Region-aware Guidance Integration.}  
The fusion process combines visual and textual cues through: 
\begin{equation}\label{eq:region-fuse}
\left\{
\begin{aligned}
  &G^{(k)} = \alpha_k\,g_{\text{image}}^{(k)} + (1-\alpha_k)\,\tilde{g}_{\text{text}}^{(k)}, \\[2pt]
  &\alpha_k = \frac{\exp\!\bigl(\lVert g_{\text{image}}^{(k)}\rVert\bigr)}
               {\exp\!\bigl(\lVert g_{\text{image}}^{(k)}\rVert\bigr)+\exp\!\bigl(\lVert \tilde{g}_{\text{text}}^{(k)}\rVert\bigr)}, \\[2pt]
  &\tilde{g}_{\text{text}}^{(k)} = W_g\!\Bigl(\sum_{n=1}^{N}\mathrm{softmax}_n(\beta\,A_{k,n})\,g_{\text{region}}[n]\Bigr), \\[2pt]
  &G_{:,i,j} = G^{(k)},\quad\forall (i,j)\in R_k,
\end{aligned}
\right.
\end{equation}
where \(G\in\mathbb{R}^{B\times H\times W\times D}\) is the region‐aware guidance tensor and \(\beta\) controls the category-wise softmax sharpness.

As illustrated in Fig.~\ref{model}, the proposed module incorporates both global semantic priors and localized region-level alignment. By combining object-aware textual refinement and region-scored visual fusion, the model reduces hallucination and enhances fine-grained category discrimination, especially in cluttered or visually ambiguous scenes.
\subsection{Contextual Cross-modal Fusion}
\label{sec:fusion}
To overcome the limitations of global feature representations in spatial precision and fine-grained alignment, we propose 
a contextual cross-modal fusion module unifying spatial detail with semantic abstraction, which is essential for accurate open-vocabulary segmentation. It integrates local structure and global semantics through three key components: first, a direction-aware attention mechanism that captures structured spatial information; second, a state-space modeling branch for long-range global context; and third, an adaptive fusion unit guided by both appearance and language cues. 

Given region-aware guidance \(G\) and CLIP middle-layer visual features \(F_{\mathrm{mid}}\in\mathbb{R}^{B\times H\times W\times C}\), we decompose both tensors evenly along channel dimensions into horizontal/vertical halves: \(G=(G_h,G_v)\), \(F_{\mathrm{mid}}=(F_{h},F_{v})\). We concatenate each pair \((F_{h}\parallel G_h)\) and \((F_{v}\parallel G_v)\) and apply rectangular self-attention—\(\mathrm{MHSA}_h\) on the horizontal branch, \(\mathrm{MHSA}_v\) on the vertical to capture directional dependencies. The two branch outputs are then concatenated:
\begin{equation}
X_{\mathrm{local}} = \bigl[\mathrm{MHSA}_h(F_{h}\parallel G_h),\, \mathrm{MHSA}_v(F_{v}\parallel G_v)\bigr].
\end{equation}

The local and global branches are then fused as:
\begin{equation}
X_{\mathrm{fused}}
= \mathcal{F}_{\mathrm{proj}}\!\bigl([\;X_{\mathrm{local}}\;\parallel\;\mathrm{SS2D}(G)\;]\bigr),
\end{equation}
where \(\mathrm{SS2D}(\cdot)\) denotes the 2D state-space global‐context module \cite{liu2024vmamba}, \(\mathcal{F}_{\mathrm{proj}}\) is a \(1\times1\) convolution for channel alignment.  

This fusion strategy enables simultaneous encoding of fine-grained spatial structures and long-range semantic relationships, significantly enhancing cross-modal representation quality. By adaptively combining local and global cues, the model achieves superior pixel-text alignment and region-level discrimination, particularly in complex scenes.

Channel adaptation employs a sequential structure consisting of a $1\times1$ convolution, a depth‐wise convolution, a gated linear unit block, and a second $1\times1$ projection, each followed by a residual connection and layer normalization. For text integration, we introduce $T$ learnable query embeddings $Q_{\mathrm{lrn}}\in\mathbb{R}^{B\times T\times D}$ and pooled text embeddings $E_{\mathrm{text}}\in\mathbb{R}^{B\times L_t\times D}$, where text features are linearly projected to dimension $D$. We then form the key and value matrices as follows: 
\begin{equation}
K = \bigl[X_{\mathrm{fused}}\parallel E_{\mathrm{text}}\bigr],\quad
V = \bigl[X_{\mathrm{fused}}\parallel E_{\mathrm{text}}\bigr],
\end{equation}
the linear attention \(\mathrm{Attn}(Q_{\mathrm{lrn}},K,V)\) is adopted, followed by token-wise gated fusion and a residual MLP.

In summary, our fusion module hierarchically integrates local patterns, global context, and language semantics. This approach enables precise cross-modal alignment and improves region-level discrimination in complex scenes.


\subsection{Decoder and Loss Function}
\label{sec:decoder}
\paragraph{Decoder.}  
We construct a correlation tensor $F\in\mathbb{R}^{B\times C\times T\times H\times W}$ by interacting $T$ learnable queries with the fused feature map. The decoding procedure is as follows:
\begin{equation}
F_{i+1} = \text{Decoder}_i\big(\text{Upsample}(F_i),\, S_i\big),
\end{equation}
where for $i{=}1,2$, $S_i$ denotes multi-scale \emph{spatial} guidance features extracted from intermediate CLIP encoder activations, resized to match the decoder's resolution. Each decoding stage performs learnable upsampling followed by a guidance-modulated convolution, enabling precise reconstruction of fine-grained structures.

Finally, the decoder output is projected to per-class logits and then upsampled to the original input resolution $(H_0,W_0)$ via bilinear interpolation:
\begin{equation}
\hat{Y}_{\text{final}} = \text{Interp}(\hat{Y}[:, :H, :W], \text{size}=(H_0,W_0)).
\end{equation}


\paragraph{Loss Function.}  
A pixel-wise multi-label binary cross-entropy loss is employed,
\begin{equation}
\begin{aligned}
\mathcal{L}_{\mathrm{BCE}}
&= -\frac{1}{\sum_{i,j}M_{i,j}}
   \sum_{i,j,n}M_{i,j}\Bigl[Y_{i,j,n}\log\sigma(\hat{Y}_{i,j,n}) \\[-2pt]
&\quad\; +(1-Y_{i,j,n})\log\bigl(1-\sigma(\hat{Y}_{i,j,n})\bigr)\Bigr],
\end{aligned}
\end{equation}
where  \(\hat{Y}_{i,j,n}\) denotes the predicted logits, \(Y_{i,j,n}\in\{0,1\}\) represent the ground-truth labels and mask \(M_{i,j}\in\{0,1\}\),  \(\sigma(x)=1/(1+e^{-x})\) is the sigmoid function.

\input{tables/main_table}

\section{Experiments}
\subsection{Datasets and Evaluation Metric}
Our experiments adhere to established OVSS protocols \cite{ghiasi2022scaling, liang2023open, cho2024cat, xie2024sed}. The model is trained on COCO-Stuff \cite{caesar2018coco}, which comprises 118K images annotated with 171 semantic categories. For comprehensive evaluation, we test on three widely-used benchmarks: ADE20K \cite{zhou2019semantic}, PASCAL VOC \cite{everingham2010pascal}, and PASCAL-Context \cite{mottaghi2014role}.
\paragraph{ADE20K.} The dataset provides 20 K training and 2 K validation images. We evaluate two splits: A-150, comprising the 150 most frequent categories, and A-847, which includes 847 categories following the split of \cite{ding2022decoupling}. 
\paragraph{PASCAL VOC.} The dataset contains 1.5 K training and 1.5 K validation images annotated with 20 foreground object classes and a background label. We report results for the PAS-20 setting (foreground classes only) and for PAS-20\textsuperscript{b}, where the background is redefined to include categories from PC-59 that are not in PAS-20, as in \cite{ghiasi2022scaling}. 
\paragraph{PASCAL-Context.} This extension of PASCAL VOC offers dense pixel-level annotations. We follow the standard PC-59 and PC-459 splits, covering 59 and 459 semantic categories, respectively, to evaluate both common and fine-grained concepts.

Results are reported as the mean Intersection-over-Union (mIoU), a metric that averages IoU over all categories and is standard in segmentation benchmarks. Higher mIoU indicates better performance.

\subsection{Implementation Details}
Our framework leverages CLIP's Vision Transformer-based image encoder while keeping the text encoder frozen during training. 
All experiments were conducted on a $4\times$ NVIDIA RTX 4090 GPU setup with a total batch size of 4 across roughly 80K training iterations.

We employ the AdamW \cite{loshchilov2017decoupled} optimizer with an initial learning rate of $2 \times 10^{-4}$ and a weight decay of $1 \times 10^{-4}$. To stabilize training, the learning rate of the image encoder is scaled by a factor of 0.01. A warm-up cosine learning rate schedule is used to gradually adjust the learning rate over time.

Training images are first cropped and resized during preprocessing to match the input resolution requirements of the CLIP backbone. To improve model robustness, we apply standard data augmentation techniques, including random horizontal flipping during training. During inference, images are uniformly resized and segmented using a sliding-window strategy to ensure full spatial coverage. For text guidance, we construct category prompts using the template \textit{``A photo of a \{category\} in the scene''}. All experiments are implemented in PyTorch \cite{paszke2019pytorch} and built upon the Detectron2 framework \cite{wu2019detectron2}.

\subsection{Comparison With State-of-the-art Methods}
We extensively evaluate our proposed LoGoSeg against leading OVSS methods on six standard benchmarks—A-847, PC-459, A-150, PC-59, PAS-20, and PAS-20\textsuperscript{b} (see Table~\ref{tab:ref_0_MainTable}). Our evaluation groups methods by type and scale of their vision-language models and provides detailed annotations on training data, external datasets, and supplementary feature backbones. The best and second-best performances are highlighted in bold and underlined.

Most approaches rely on ViT backbones and often use extra data (e.g., COCO Captions \cite{Chen2015MicrosoftCC}, YFCC \cite{thomee2016yfcc100m}) or additional feature extractors (e.g., ResNet-101 \cite{he2016deep}, Swin-B \cite{liu2021swin}). In contrast, LoGoSeg employs a unified VLM backbone without external datasets or  auxiliary modules. Moreover, our updated design supports diverse backbone configurations beyond the original ViT-B/16 and ViT-L/14 settings.

\paragraph{Backbone Scaling.}  Table~\ref{tab:ref_0_MainTable} also reveals the scalability of different methods when transitioning from medium-size (e.g., ViT-B/16, ConvNeXt-B) to large backbones (ViT-L/14, ConvNeXt-L). Most two-stage frameworks show marginal gains, whereas LoGoSeg achieves near-linear improvements, consistently ranking first or second on  all benchmarks (after scaling). This confirms the robustness of our object prior, region alignment, and dual-stream fusion across model capacities. 

\paragraph{Backbone Configurations and Scaling.} Using a medium backbone (ViT-B/16), LoGoSeg reaches 12.6 mIoU on A-847 and 19.5 on PC-459. When switching to the larger ViT-L/14 backbone, performance consistently improves, achieving 16.4 on A-847, 24.5 on PC-459, 38.2 on A-150, 63.6 on PC-59, 97.4 on PAS-20, and 83.3 on PAS-20\textsuperscript{b}. These results indicate a clear and consistent scaling trend. ConvNeXt-B and ConvNeXt-L variants \cite{liu2022convnet} exhibit similar behavior and match or surpass their ViT counterparts. All configurations operate without external data or auxiliary modules, confirming that increasing backbone capacity translates almost linearly into accuracy gains for LoGoSeg.

\paragraph{One-stage vs. Two-stage Frameworks.} Two-stage models like ZegFormer, OVSeg, FC-CLIP, and MAFT+ separate mask generation and classification, whereas one-stage models (SAN, CAT-Seg, SED, EBSeg) predict masks directly but emphasise either local or global cues. LoGoSeg unifies both cues in a single stage, achieving the best or second-best mIoU on all six benchmarks, as shown in Table~\ref{tab:ref_0_MainTable}. It outperforms the strongest two-stage competitor (MAFT+) by 4.2 mIoU on PC-59 and the best one-stage baseline (SED) by 1.9 mIoU on PC-459. LoGoSeg is the only one-stage method that leading on both ADE20K splits (A-150 and A-847), highlighting its balanced capability for common and long-tail categories. These results confirm that our object prior, region-aware alignment, and dual-stream fusion remain effective across backbone scales and framework styles.

\begin{figure*}[t]
\centering
\includegraphics[width=1.0\linewidth]{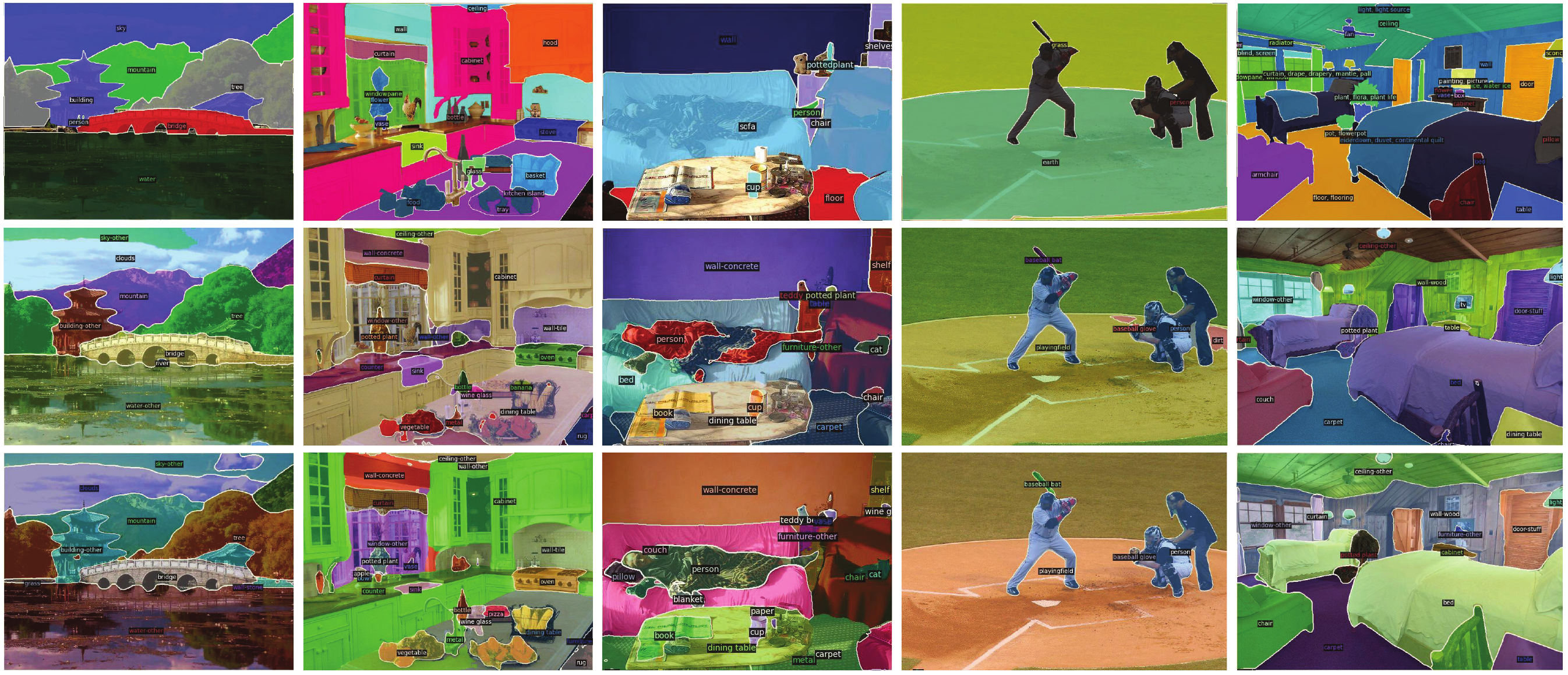}
\caption{\textbf{Qualitative comparison.} Rows show ground truth, SED predictions, and LoGoSeg results. Region-aware alignment and dual-branch fusion yield more stable, less hallucinatory segmentations. LoGoSeg correctly identifies \textit{pizza} and \textit{cat} missing from the ground truth and avoids SED’s mislabels (e.g., vegetables as \textit{banana}, paintings as \textit{TV}).}
\label{result}
\end{figure*}

Fig.~\ref{result} shows qualitative comparisons. The first row presents ground truth annotations, the second row displays SED predictions, and the third row shows LoGoSeg results. Thanks to region-aware alignment and dual-branch fusion, our method yields more stable segmentation with fewer hallucinations, correctly identifying objects like \textit{pizza} (col. 2) and \textit{cat} (col. 3), while avoiding misclassifications by SED. Failure cases reveal that LoGoSeg may under-segment small or heavily occluded objects (e.g., the wine glass in col. 2 and the potted plant in col. 5) and confuse visually similar classes under low-contrast conditions. See the Appendix for additional qualitative results.

Overall, LoGoSeg demonstrates robust performance across diverse segmentation benchmarks without relying on additional data, handcrafted prompts, or auxiliary modules.

\input{tables/Table2}
\input{tables/Table3}

\subsection{Ablation Study}

We conduct ablation studies to evaluate the effectiveness of each module in  LoGoSeg. All experiments are conducted under ViT-B/16 unless stated otherwise.

\paragraph{Impact of Core Modules.} Table \ref{tab:core_modules_combination} shows the impact of integrating Object Prior (OP), Contextual Cross-modal Fusion (CCF), and Region Alignment (RA). The baseline uses global features for pixel-level segmentation without explicit local or regional alignment, yielding relatively lower scores. Incorporating OP notably reduces classification hallucinations, improving performance on all benchmarks. Adding CCF further enhances local-global feature integration, resulting in consistent performance gains. Integrating RA provides the most substantial improvement, demonstrating its effectiveness in enhancing pixel-text alignment.

\paragraph{Contextual Cross-modal Fusion Analysis.} Table~\ref{tab:fusion_ablation} evaluates the contributions of individual components in the fusion module. LangQuery alone already improves performance, indicating the effectiveness of basic text-guided attention. Incorporating GlobalContext further enhances accuracy across all benchmarks, highlighting the role of long-range semantic modeling. Directional attention brings additional gains, particularly on fine-grained datasets, due to improved sensitivity to spatial structure. Analysis of fusion depth shows that a two-layer design achieves a favorable balance between performance and complexity. Overall, these results underscore the importance of jointly modeling spatial and semantic information.

\input{tables/Table4}

\paragraph{Prior-Guided Alignment Hyperparameters.} We examine the effects of two hyperparameters in Table~\ref{tab:alignment_hyper}: the semantic prior weight (\(\lambda\)) and region size (\(r \times r\) grid). Results show that region alignment without object priors (\(\lambda=0\)) achieves limited performance, while prior guidance yields significant improvements (optimal at \(\lambda=1\)). For region granularity, a \(6 \times 6\) partitioning provides a balance between localized detail preservation and computational efficiency, outperforming both larger and smaller grids.

\input{tables/Table6}

\paragraph{Training Dataset Generalization.} Table~\ref{tab:scalability} evaluates LoGoSeg under three regimes: full COCO-Stuff, PC-59 (59 categories), and A-150 (150 categories). With PC-59 supervision, LoGoSeg lifts A-847 mIoU from 3.8 to 10.1 (+6.3) and PC-459 from 8.2 to 17.3 (+9.1) over ZegFormer, demonstrating strong transfer. Under A-150 supervision, it attains 15.5 mIoU on A-847 and 16.7 on PC-459, surpassing CAT-Seg by 2.2 points. These results verify that region-aware alignment and dual-branch fusion maintain stable performance across varying annotation scales and datasets. 
\section{Conclusion}
We present LoGoSeg, a single-stage framework for OVSS that integrates object priors, region-level alignment, and local-global feature fusion. Unlike existing approaches, our method establishes region-text correspondences while suppressing irrelevant categories and addressing prediction hallucination. The dual-stream fusion architecture captures fine-grained spatial structures and global semantic context. Without external datasets or auxiliary backbones, LoGoSeg generalizes across six benchmarks. It achieves strong performance, but its reliance on large vision-language backbones may constrain real-time deployment in resource-limited scenarios. Future work includes developing lightweight variants, incorporating advanced visual encoders, and supporting multimodal inputs beyond vision-language pairs to enhance scalability and applicability.

\section*{Acknowledgments}
This work was supported in part by the National Natural Science Foundation of China under Grants 62306072 and 62576093, and in part by the Fundamental Research Funds for the Central Universities under Grant 2242025K30024.


\bibliography{aaai2026}
\end{document}

%% file: tables/main_table.tex
\begin{table*}[htp]
\begin{center}
\resizebox{\textwidth}{!}{
\Large
\renewcommand{\arraystretch}{1.2}

\begin{tabular}{lcccccccccc}
\toprule
Method & VLM & Feature backbone & Training Dataset & Additional Dataset & \texttt{A-847} & \texttt{PC-459} & \texttt{A-150} & \texttt{PC-59} & \texttt{PAS-20} & \texttt{PAS-20$^b$} \\
\midrule
GroupViT \cite{xu2022groupvit} & ViT-S/16 & - & GCC+YFCC & \cmark & 4.3 & 4.9 & 10.6 & 25.9 & 50.7 & - \\
ZegFormer \cite{ding2022decoupling} & ViT-B/16 & ResNet-101 & COCO-Stuff-156 & \xmark & 4.9 & 9.1 & 16.9 & 42.8 & 86.2 & 62.7 \\
ZSseg \cite{xu2022simple} & ViT-B/16 & ResNet-101 & COCO-Stuff & \xmark & 7.0 & - & 20.5 & 47.7 & 88.4 & - \\
OpenSeg \cite{ghiasi2022scaling} & ALIGN & ResNet-101 & COCO Panoptic & \cmark & 4.4 & 7.9 & 17.5 & 40.1 & - & 63.8 \\
DeOP \cite{han2023open} & ViT-B/16 & ResNet-101c & COCO-Stuff-156 & \xmark & 7.1 & 9.4 & 22.9 & 48.8 & 91.7 & - \\
PACL \cite{mukhoti2023open} & ViT-B/16 & - & GCC+YFCC & \cmark & - & - & 31.4 & 50.1 & 72.3 & - \\
OVSeg \cite{liang2023open} & ViT-B/16 & ResNet-101c & COCO-Stuff+COCO Caption & \cmark & 7.1 & 11.0 & 24.8 & 53.3 & 92.6 & - \\
SAN \cite{xu2023san} & ViT-B/14 & - & COCO-Stuff & \xmark & 10.1 & 12.6 & 27.5 & 53.8 & 94.0 & - \\
CAT-Seg \cite{cho2024cat} & ViT-B/16 & ResNet-101 & COCO-Stuff & \xmark & 8.4 & 16.6 & 27.2 & 57.5 & 93.7 & \underline{78.3} \\
EBSeg \cite{shan2024open} & ViT-B/16 & - & COCO-Stuff & \xmark & 11.1 & 17.3 & 30.0 & 56.7 & 94.6 & - \\
SED \cite{xie2024sed} & ConvNeXt-B & - & COCO-Stuff & \xmark & 11.4 & 18.6 & 31.6 & 57.3 & 94.4 & - \\

\midrule

 & ConvNeXt-B & - & COCO-Stuff & \xmark & \underline{12.2} & \underline{19.3} & \underline{32.0} & \underline{57.8} & \underline{95.0} & 77.2 \\

\multirow{-2}{*}{\textbf{LoGoSeg (ours)}}  
 & ViT-B/16 & - & COCO-Stuff & \xmark & \textbf{12.6} & \textbf{19.5} & \textbf{32.6} & \textbf{58.2} & \textbf{95.4} & \textbf{78.6} \\

\midrule
LSeg \cite{lseg} & ViT-B/32 & ViT-L/16 & PASCAL VOC-15 & \xmark & - & - & - & - & 52.3 & - \\
OpenSeg \cite{ghiasi2022scaling} & ALIGN & Eff-B7 & COCO Panoptic & \cmark & 8.1 & 11.5 & 26.4 & 44.8 & - & 70.2 \\
OVSeg \cite{liang2023open} & ViT-L/14 & Swin-B & COCO-Stuff+COCO Caption & \cmark & 9.0 & 12.4 & 29.6 & 55.7 & 94.5 & - \\
ODISE \cite{xu2023open} & ViT-L/14 & - & COCO Panoptic & \cmark & 11.1 & 14.5 & 29.9 & 57.3 & - & - \\
HIPIE \cite{wang2023hierarchical} & BERT-B & ViT-H & COCO Panoptic & \cmark & - & - & 29.0 & 59.3 & - & - \\
SAN \cite{xu2023san} & ViT-L/14 & - & COCO-Stuff & \xmark & 13.7 & 17.1 & 33.3 & 60.2 & 95.5 & - \\
CAT-Seg \cite{cho2024cat} & ViT-L/14 & Swin-B & COCO-Stuff & \xmark & 10.8 & 20.4 & 31.5 & 62.0 & 96.6 & 81.8 \\
FC-CLIP \cite{yu2023convolutions} & ConvNeXt-L & - & COCO Panoptic & \cmark & 14.8 & 18.2 & 34.1 & 58.4 & 95.4 & - \\
EBSeg \cite{shan2024open} & ViT-L/14 & - & COCO-Stuff & \xmark & 13.7 & 21.0 & 32.8 & 60.2 & 96.4 & - \\
SED \cite{xie2024sed} & ConvNeXt-L& - & COCO-Stuff & \xmark & 13.9 & 22.6 & 35.2 & 60.6 & 96.1 & - \\
MAFT+ \cite{jiao2024collaborative} & ConvNeXt-L & - & COCO-Stuff & \xmark & 15.1 & 21.6 & 36.1 & 59.4 & 96.5 & - \\
\midrule

 & ConvNeXt-L & - & COCO-Stuff & \xmark & \underline{15.8} & \underline{23.6} & \underline{37.8} & \underline{63.2} & \underline{97.0} & \underline{82.1} \\

\multirow{-2}{*}{\textbf{LoGoSeg (ours)}}  
 & ViT-L/14 & - & COCO-Stuff & \xmark & \textbf{16.4} & \textbf{24.5} & \textbf{38.2} & \textbf{63.6} & \textbf{97.4} & \textbf{83.3} \\


\bottomrule
\end{tabular}
}
\caption{\textbf{Comparison with state-of-the-art methods.} 
OVSS performance on six benchmarks, reported in mIoU (\%). 
Bold entries denote the best results, and underlined entries indicate the second best.}
\label{tab:ref_0_MainTable}

\end{center}
\end{table*}

%% file: tables/Table2.tex
\begin{table}[t]
    \centering
    \resizebox{\linewidth}{!}{
    \begin{tabular}{ccc|cccccc}
        \toprule
        OP & CCF & RA & A-847 & PC-459 & A-150 & PC-59 & PAS-20 & $\textnormal{PAS-20}^b$ \\
        \midrule
         &  &  & 10.8 & 17.6 & 28.6 & 53.9 & 94.3 & 73.1 \\
        \cmark &  &  & 11.5 & 18.1 & 29.4 & 56.4 & 94.6 & 75.8 \\
        \cmark & \cmark &  & 12.3 & 19.0 & 32.0 & 57.8 & 94.8 & 76.9 \\
         \cmark & \cmark & \cmark & \textbf{12.6} & \textbf{19.5} & \textbf{32.6} & \textbf{58.2} & \textbf{95.4} & \textbf{78.6} \\
        \bottomrule
    \end{tabular}
    }
        \caption{\textbf{Ablation study of core modules.}
Object Prior (OP), Contextual Crossmodal Fusion (CCF), and Region Alignment (RA) are evaluated using CLIP (ViT-B/16). Results are in mIoU (\%).}

    \label{tab:core_modules_combination}

\end{table}

%% file: tables/Table3.tex
\begin{table}[t]
  \centering
  \resizebox{\linewidth}{!}{
    \begin{tabular}{l|cccccc}
      \toprule
      Method                   & A-847 & PC-459 & A-150 & PC-59 & PAS-20 & PAS-20$^b$ \\
      \midrule
      w/ LQ only               & 10.9  & 17.8   & 29.4  & 55.5  & 93.2   & 75.3      \\
      w/ LQ + GC               & 11.8  & 18.4   & 31.2  & 56.7  & 94.6   & 76.8      \\
      w/ LQ + GC + DA          & \textbf{12.6} & \textbf{19.5} & \textbf{32.6} & \textbf{58.2} & \textbf{95.4} & \textbf{78.6} \\  
      \midrule
      w/ fusion depth = 1      & 11.5  & 18.5   & 31.6  & 56.9  & 94.4   & 76.6      \\
      w/ fusion depth = 2      & \textbf{12.6} & \textbf{19.5} & \textbf{32.6} & \textbf{58.2} & \textbf{95.4} & \textbf{78.6} \\
      w/ fusion depth = 3      & 11.9  & 18.8   & 31.5  & 57.4  & 94.8   & 77.2      \\
      \bottomrule
    \end{tabular}
  }
  \caption{\textbf{Ablation of Contextual Crossmodal Fusion.} We analyze the contribution of LangQuery (LQ), GlobalContext (GC), and Dir-Attn (DA), together with different fusion depths. Results are shown in mIoU (\%).}
  \label{tab:fusion_ablation}
\end{table}

%% file: tables/Table4.tex
\begin{table}[t]
    \centering
    \resizebox{\linewidth}{!}{
    \begin{tabular}{cc|cccccc}
        \toprule
        $\lambda$ & $r \times r$ & A-847 & PC-459 & A-150 & PC-59 & PAS-20 & $\textnormal{PAS-20}^b$ \\
        \midrule
        0 & 4$\times$4 & 11.6 & 18.8 & 31.6 & 56.9& 94.5 & 76.4\\
        0 & 6$\times$6 & 11.8& 18.5 & 31.8 & 57.5& 94.8 & 76.9\\
        0 & 8$\times$8 & 11.5 & 18.5 & 31.4 & 56.6 & 94.9 & 75.8 \\
        0.5 & 6$\times$6 & 12.0 & 18.9 & 31.9 & 57.5 & 95.2& 77.4\\
        0.5 & 8$\times$8 & 11.4& 18.3& 31.2& 57.3& 94.8& 77.0\\
        Adapt. & 6$\times$6 & 12.3 & 18.9 & 32.3 & 57.6 & 95.0 & 76.8 \\
        Full & 6$\times$6 & \textbf{12.6} & \textbf{19.5} & \textbf{32.6} & \textbf{58.2} & \textbf{95.4} & \textbf{78.6} \\
        \bottomrule
    \end{tabular}
    }
    \caption{\textbf{Ablation of Prior-Guided Alignment Hyperparameters.}
Varying $\lambda$ and region size $r\times r$, we report mIoU (\%). 
The \textit{Full} setting ($\lambda=1$, $6\times6$ grid) performs best. 
“Adapt.” denotes a learnable parameter.}

    \label{tab:alignment_hyper}

\end{table}

%% file: tables/Table6.tex
\begin{table}[!htbp]
    \centering
    \large
    \renewcommand{\arraystretch}{1.2}
    \resizebox{\linewidth}{!}{
    \begin{tabular}{l|c|cccccc}
        \toprule
        Methods & Training dataset & A-847 & PC-459 & A-150 & PC-59 & PAS-20 & $\textnormal{PAS-20}^b$ \\
        \midrule
        ZegFormer & COCO-Stuff & 5.6 & 10.4 & 18.0 & 45.5 & 89.5 & 65.5 \\
        ZSseg  & COCO-Stuff & 7.0 & 9.0 & 20.5 & 47.7 & 88.4 & 67.9 \\
        CAT-Seg  & COCO-Stuff & 8.4 & 16.6 & 27.2 & 57.5 & 93.7 & 78.3 \\
        LoGoSeg (Ours) & COCO-Stuff & \textbf{12.6} & \textbf{19.5} & \textbf{32.6} & \textbf{58.2} & \textbf{95.4} & \textbf{78.6} \\
        \midrule
        
        ZegFormer  & PC-59 & 3.8& 8.2& 13.1& \textcolor{gray}{48.7}& 86.5& 66.8\\
        ZSseg  & PC-59 & 3.0& 7.6& 11.9& \textcolor{gray}{54.7}& 87.7& 71.7\\
        CAT-Seg  & PC-59 & 5.6& 12.9& 23.0& \textcolor{gray}{62.4}& 87.3& 79.0\\
        LoGoSeg (Ours) & PC-59 & \textbf{10.1} & \textbf{17.3} & \textbf{28.2} & \textcolor{gray}{63.6} & \textbf{93.1} & \textbf{80.1} \\
        
        \midrule
        ZegFormer  & A-150 & 6.8 & 7.1 & \textcolor{gray}{33.1} & 34.7 & 77.2 & 53.6 \\
        ZSseg  & A-150 &  7.6& 7.1 & \textcolor{gray}{40.3}&  39.7& 80.9& 61.1\\
        CAT-Seg  & A-150 & 10.6 & 14.5 & \textcolor{gray}{46.8}& 46.7& 85.5& 70.3\\
        LoGoSeg (Ours) & A-150 & \textbf{15.5} & \textbf{16.7} &\textcolor{gray}{47.6} & \textbf{50.8} & \textbf{91.9} & \textbf{72.7} \\
        \midrule

    \end{tabular}
    }
    \caption{\textbf{Training Dataset Generalization.}
Using CLIP (ViT-B/16), we compare mIoU (\%) across training sets. 
LoGoSeg maintains strong generalization even with limited data. 
Training-set scores are shown in \textcolor{gray}{gray}.}

    \label{tab:scalability}

\end{table}